%% file: main.tex
\documentclass[10pt,twocolumn,letterpaper]{article}

\usepackage{cvpr}              
\usepackage{multirow}
\usepackage{wrapfig}
\usepackage{adjustbox}

\input{preamble}

\definecolor{cvprblue}{rgb}{0.21,0.49,0.74}
\usepackage[pagebackref,breaklinks,colorlinks,citecolor=cvprblue]{hyperref}

\def\paperID{*****} 
\def\confName{CVPR}
\def\confYear{2024}

\title{SAM2-Adapter: Evaluating \& Adapting Segment Anything 2 in Downstream Tasks: Camouflage, Shadow, Medical Image Segmentation, and More}

\author{Tianrun Chen$^{1,2 +*}$\and Ankang Lu$^{3 +}$\and Lanyun Zhu$^{4 +}$\and Chaotao Ding$^{1+}$ \\ \and Chunan Yu $^{3}$ \and Deyi Ji $^{6}$ \and Zejian Li $^{5}$ \and Lingyun Sun $^{2}$ \and Papa Mao$^{1}$ \\\and Ying Zang$^{3*}$\\
\\
\textbf{This Work is Built Upon the \href{http://tianrun-chen.github.io/SAM-Adaptor/}{SAM-Adapter - First Available at 14 April, 2023}} \\
\textbf{TL; DR}. SAM2, enhanced with our new adapter, can replace SAM as the backbone for segmentation tasks, \\\textbf{establishing new state-of-the-art (SOTA) results for downstream tasks!}\\
\\
\small $^+$ Equal Contribution   $^*$ Corresponding Author \\\small$\{$tianrun.chen@zju.edu.cn; 02750@zjhu.edu.cn$\}$
\\
\small $^{1}$KOKONI, Moxin (Huzhou) Tech. Co., LTD, Huzhou, Zhejiang, P.R. China.\\ 
\small $^{2}$College of Computer Science and Technology, Zhejiang University, Hangzhou, Zhejiang, P.R. China.\\ 
\small $^{3}$School of Information Engineering, Huzhou University, Huzhou, Zhejiang, P.R. China.\\ 
\small $^{4}$ Information Systems Technology
and Design Pillar, Singapore University of Technology and Design, Singapore.\\
\small $^{5}$ School of Software Technology, Zhejiang University, Hangzhou, Zhejiang, P.R. China.\\ 
\small $^{6}$ School of Information Science and Technology, University of Science and Technology of China, P.R. China.\\ 
\\
Project Page: \href{http://tianrun-chen.github.io/SAM-Adaptor/}{http://tianrun-chen.github.io/SAM-Adaptor/}
}

\begin{document}
\maketitle
\input{sec/0_abstract}    
\input{sec/1_intro}
\input{sec/2_Related_Work}

\input{sec/3_Method}
\input{sec/4_Experiments}
\input{sec/5_Conclusion}

{
    \small
    \bibliographystyle{ieeenat_fullname}
    \bibliography{main}
}

\end{document}

%% file: preamble.tex
\usepackage[dvipsnames]{xcolor}


%% file: sec/0_abstract.tex
\begin{abstract}
    The advent of large models, also known as foundation models, has significantly transformed the AI research landscape, with models like Segment Anything (SAM) achieving notable success in diverse image segmentation scenarios. Despite its advancements, SAM encountered limitations in handling some complex low-level segmentation tasks like camouflaged object and medical imaging. In response, in 2023, we introduced SAM-Adapter, which demonstrated improved performance on these challenging tasks. Now, with the release of Segment Anything 2 (SAM2)—a successor with enhanced architecture and a larger training corpus—we reassess these challenges. This paper introduces SAM2-Adapter, the first adapter designed to overcome the persistent limitations observed in SAM2 and achieve new state-of-the-art (SOTA) results in specific downstream tasks including medical image segmentation, camouflaged (concealed) object detection, and shadow detection. SAM2-Adapter builds on the SAM-Adapter's strengths, offering enhanced generalizability and composability for diverse applications. We present extensive experimental results demonstrating SAM2-Adapter's effectiveness. We show the potential and encourage the research community to leverage the SAM2 model with our SAM2-Adapter for achieving superior segmentation outcomes. Code, pre-trained models, and data processing protocols are available at \href{http://tianrun-chen.github.io/SAM-Adaptor/}{http://tianrun-chen.github.io/SAM-Adaptor/}
\end{abstract}

%% file: sec/1_intro.tex
\section{Introduction}
\label{sec:intro}
     The AI research landscape has been transformed by foundation models trained on vast data \cite{bommasani2021opportunities, zhu2024llafs, zhu2024ibd, chen2024reasoning3d}.  Recently, among the foundation models, Among these, Segment Anything (SAM) \cite{SAM} stands out as a highly successful image segmentation model with demonstrated success in diverse scenarios. However, in our previously study, we found that SAM's performance was limited in some challenging low-level structural segmentation tasks, such as camouflaged object detection and shadow detection. To address this, in 2023, within two weeks of SAM's release, we proposed the SAM-Adapter \cite{chen2023samfailssegmentanything, chen2023sam}, which aimed to leverage the power of the SAM model to deliver better performance on these challenging downstream tasks. The success of the SAM-Adapter, with its training and evaluation code and checkpoints made publicly available, has already been a valuable resource for many researchers in the community to experiment with and build upon, demonstrating its effectiveness on a variety of downstream tasks.
    
     Now, the research community has pushed the boundaries further with the introduction of an even more capable and versatile successor to SAM, known as Segment Anything 2 (SAM2). Boasting further enhancements in its network architecture and training on an even larger visual corpus, SAM2 has certainly piqued our interest. This naturally leads us to the questions: 
     
 \begin{itemize}
     \item Do the challenges faced by SAM in downstream tasks persist in SAM2?
     \item Can we replicate the success of SAM-Adapter and leverage SAM2's more powerful pre-trained encoder and decoder to achieve new state-of-the-art (SOTA) results on these tasks?
 \end{itemize} 

     In this paper, we answer both questions with a resounding "Yes."  Our experiments confirm that the challenges SAM encountered in downstream tasks do persist in SAM2, due to the inherent limitations of foundation models—where training data cannot cover the entire corpus and working scenarios vary \cite{bommasani2021opportunities}. However, we have devised a solution to address this challenge. By introducing the \textbf{SAM2-Adapter}, we’ve created a multi-adapter configuration that leverages SAM2's enhanced components to achieve new SOTA results in tasks including medical image segmentation, camouflaged object detection, and shadow detection.
      
     Just like SAM-Adapter \cite{chen2023samfailssegmentanything, chen2023sam}, \textbf{this pioneering work is the first attempt to adapt the large pre-trained segmentation model SAM2 to specific downstream tasks and achieve new SOTA performance}. SAM2-Adapter builds on the strengths of the original SAM-Adapter while introducing significant advancements.

    SAM2-Adapter inherits the core advantages of SAM-Adapter, including:
    
\begin{itemize}
\item \textbf{Generalizability}: SAM2-Adapter can be directly applied to customized datasets of various tasks, enhancing performance with minimal additional data. This flexibility ensures that the model can adapt to a wide range of applications, from medical imaging to environmental monitoring.

\item \textbf{Composability}: SAM2-Adapter supports the easy integration of multiple conditions to fine-tune SAM2, improving task-specific outcomes. This composability allows for the combination of different adaptation strategies to meet the specific requirements of diverse downstream tasks.
 \end{itemize}

     SAM2-Adapter enhances these benefits by adapting to SAM2's multi-resolution hierarchical Transformer architecture. By employing multiple adapters working in tandem, SAM2-Adapter effectively leverages SAM2’s multi-resolution and hierarchical features for more precise and robust segmentation, which
     maximizes the potential of the already-powerful SAM2. We perform extensive experiments on multiple tasks and datasets, including ISTD for shadow detection \cite{wang2018stacked} and COD10K \cite{fan2020camouflaged}, CHAMELEON \cite{skurowski2018animal}, CAMO \cite{le2019anabranch} for camouflaged object detection task, and kvasir-SEG \cite{jha2020kvasir} for polyp segmentation (medical image segmentation) task. Benefiting from the capability of SAM2 and our SAM-Adapter, our method achieves state-of-the-art (SOTA) performance on both tasks. The contributions of this work can be summarized as follows:
 
\begin{itemize}
    \item We are the first to identify and analyze the limitations of the Segment Anything 2 (SAM2) model in specific downstream tasks, continuing our research from SAM.
    \item Second, we are the first to propose the adaptation approach, \textbf{SAM2-Adapter}, to adapt SAM2 to downstream tasks and achieve enhanced performance. This method effectively integrates task-specific knowledge with the general knowledge learned by the large model.
    \item Third, despite SAM2's backbone being a simple plain model lacking specialized structures tailored for the specific downstream tasks, our extensive experiments demonstrate that SAM2-Adapter achieves SOTA results on challenging segmentation tasks, setting new benchmarks and proving its effectiveness in diverse applications.
\end{itemize}

    By further building upon the success of the SAM-Adapter, the SAM2-Adapter inherents the advantages of SAM-Adapter and demonstrates the exceptional ability of the SAM2 model to transfer its knowledge to specific data domains, pushing the boundaries of what is possible in downstream segmentation tasks. We encourage the research community to adopt SAM2 as the backbone in conjunction with our SAM2-Adapter, to achieve even better segmentation results in various research fields and industrial applications. We are releasing our code, pre-trained model, and data processing protocols in \href{http://tianrun-chen.github.io/SAM-Adaptor/}{http://tianrun-chen.github.io/SAM-Adaptor/}.

%% file: sec/2_Related_Work.tex
\section{Related Work}
    \noindent \textbf{Semantic Segmentation.} In recent years, semantic segmentation has made significant progress, primarily due to the remarkable advancements in deep-learning-based methods such as fully convolutional networks (FCN) \cite{fcn}, encoder-decoder structures \cite{unet, bisenetv2, segnet, dlpl, chen2024xlstm, urur, sstkd}, dilated convolutions \cite{deeplabv3, deeplabv3+, liu2021label, zang2024resmatch, cdgc}, pyramid structures \cite{zhu2021learning, deeplabv3, pspnet, deeplabv3+, zhu2023continual, fu2022panoptic, cagcn, wang2023fvp}, attention modules \cite{ann, zhu2024addressing, zhu2023learning, pptformer, gpwformer}, and transformers \cite{zheng2021rethinking, xie2021segformer, strudel2021segmenter, cheng2022masked, zhu2024llafs}. Recent advancements have improved SAM's performance, such as \cite{HQ-SAM}, which introduces a High-Quality output token and trains the model on fine-grained masks. Other efforts have focused on enhancing SAM's efficiency for broader real-world and mobile use, exemplified by \cite{EfficientSAM, MobileSAM, FastSAM}. The widespread success of SAM has led to its adoption in various fields, including medical imaging \cite{Ma_2024, deng2023segmentmodelsamdigital, Mazurowski_2023, wu2023medicalsamadapteradapting, ma2024segment}, remote sensing \cite{chen2023rsprompterlearningpromptremote, ren2023segmentanythingspace, changenet}, motion segmentation \cite{xie2024movingobjectsegmentationneed, wang2021learning, ipgn, feng2018challenges}, and camouflaged object detection \cite{tang2023samsegmentanythingsam}. Notably, our previous work SAM-Adapter \cite{chen2023samfailssegmentanything, chen2023sam} tested camouflaged object detection, polyp segmentation, and shadow segmentation, and provide the first adapter-based method to integrate the SAM's exceptional capability to these downstream tasks. \

    \noindent \textbf{Adapters. }The concept of Adapters was first introduced in the NLP community \cite{houlsby2019parameter} as a tool to fine-tune a large pre-trained model for each downstream task with a compact and scalable model. In \cite{stickland2019bert}, multi-task learning was explored with a single BERT model shared among a few task-specific parameters. In the computer vision community, \cite{li2022exploring, homview, mscnn} suggested fine-tuning the ViT \cite{dosovitskiy2020image} for object detection with minimal modifications. Recently, ViT-Adapter \cite{chen2022vision} leveraged Adapters to enable a plain ViT to perform various downstream tasks. \cite{liu2023explicit} introduce an Explicit Visual Prompting (EVP) technique that can incorporate explicit visual cues to the Adapter. However, no prior work has tried to apply Adapters to leverage pretrained image segmentation model SAM trained at large image corpus. Here, we mitigate the research gap. 

\begin{figure*}[t]
\centering
\includegraphics[width=\linewidth]{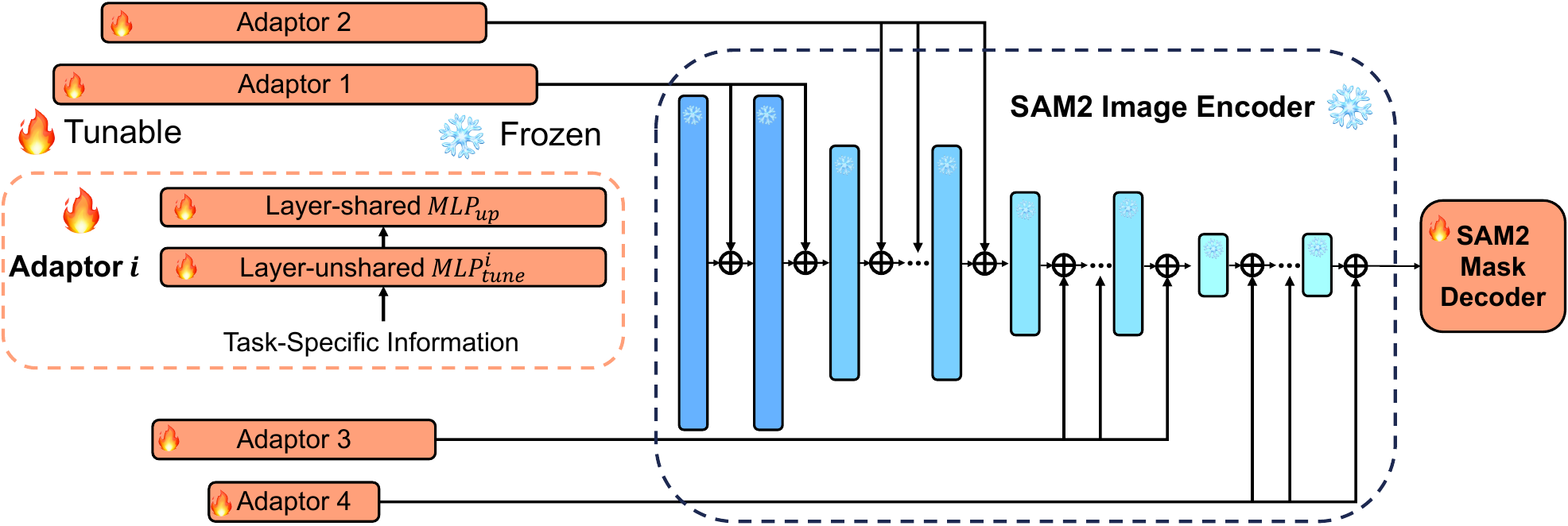}
\caption{\textbf{The architecture of the proposed SAM2-Adapter. We show the differences between SAM2-Adapter and SAM-Adapter in the Ablation Study section.} } \label{framework}
\end{figure*}

\noindent \textbf{Polyp Segmentation.}
    In recent years, there has been notable progress in polyp segmentation \cite{zhou2021fullattentionbasedneuralarchitecture} due to deep-learning approaches. These techniques employ deep neural networks to derive more discriminative features from endoscopic polyp images. Nonetheless, the use of bounding-box detectors often leads to inaccurate polyp boundary localization. To resolve this, \cite{canny1986computational} leveraged fully convolutional networks (FCN) with pre-trained models to identify and segment polyps. \cite{qadir2021toward} introduced a technique utilizing Fully Convolutional Neural Networks (FCNNs) to predict 2D Gaussian shapes. Subsequently, the U-Net \cite{kingma2017adammethodstochasticoptimization} architecture, featuring a contracting path for context capture and a symmetric expanding path for precise localization, achieved favorable segmentation results. However, these strategies focus primarily on entire polyp regions, neglecting boundary constraints. Therefore, Psi-Net \cite{murugesan2019psinetshapeboundaryaware} incorporated both region and boundary constraints for polyp segmentation, yet the interplay between regions and boundaries remained underexplored. \cite{mahmud2021polypsegnet} introduced PolypSegNet, an enhanced encoder-decoder architecture designed for the automated segmentation of polyps in colonoscopy images. To address the issue of non-equivalent images and pixels, \cite{guo2022non} proposed a confidence-aware resampling method for polyp segmentation tasks. Specifically for polyp segmentation, works done by \cite{zhou2023samsegmentpolyps} and \cite{chen2023samfailssegmentanything} present promising results using an unprompted SAM and a domain-adapted SAM respectively. Additionally, Polyp-SAM \cite{li2023polypsamtransfersampolyp} used SAM for the same task. \cite{roy2023sammdzeroshotmedicalimage} evaluated the zero-shot capabilities of SAM on the organ segmentation task.

    \noindent \textbf{Camouflaged Object Detection (COD).} Camouflaged object detection, or concealed object detection is a challenging but useful task that identifies objects that blend in with their surroundings. COD has wide applications in medicine, agriculture, and art. Initially, research of camouflage detection relied on low-level features like texture, brightness, and color \cite{feng2013camouflage,pike2018quantifying,hou2011detection,sengottuvelan2008performance} to distinguish foreground from background. It is worth noting that some of this prior knowledge is critical in identifying the objects, and is used to guide the neural network in this paper.

    Le et al.\cite{le2019anabranch} first proposed an end-to-end network consisting of a classification and a segmentation branch. Recent advances in deep learning-based methods have shown a superior ability to detect complex camouflaged objects \cite{fan2020camouflaged, mei2021camouflaged, lin2023frequency}. In this work, we leverage the advanced neural network backbone (a foundation model -- SAM2) with the input of task-specific prior knowledge to achieve state-of-the-art (SOTA) performance. 

    \noindent \textbf{Shadow Detection.} Shadows can occur when an object's surface is not directly exposed to light. They offer hints on light source direction and scene illumination that can aid scene comprehension \cite{karsch2011rendering,lalonde2012estimating}. They can also negatively impact the performance of computer vision tasks \cite{nadimi2004physical,cucchiara2003detecting}. Early methods use hand-crafted heuristic cues like chromaticity, intensity, and texture \cite{huang2011characterizes,lalonde2012estimating,zhu2010learning}. Deep learning approaches leverage the knowledge learned from data and use delicately designed neural network structures to capture the information (e.g. learned attention modules) \cite{le2018a+,cun2020towards,zhu2018bidirectional}. This work leverages the heuristic priors with large neural network models to achieve the state-of-the-art (SOTA) performance.

%% file: sec/3_Method.tex
\section{Method}

\subsection{Using SAM 2 as the Backbone}
The core of our SAM2-Adapter is built upon the powerful image encoder and mask decoder components of the SAM2 model. Specifically, we leverage the MAE pre-trained Hiera image encoder from SAM2, keeping its weights frozen to preserve the rich visual representations it has learned from pretraining on large-scale datasets. Additionally, we utilize the mask decoder module from the original SAM2 model, initializing its weights with the pretrained SAM2 parameters and then fine-tuning it during the training of our adapter. We do not provide any additional prompts as input to the original SAM2 mask decoder. 

Similar to the successful approach of the SAM-Adapter \cite{chen2023samfailssegmentanything}, we next learn and inject task-specific knowledge $F^i$ into the network via Adapters. We employ the concept of prompting, which utilizes the fact that foundation models like SAM2 have been trained on large-scale datasets. Using appropriate prompts to introduce task-specific knowledge \cite{liu2023explicit} can enhance the model's generalization ability on downstream tasks, especially when annotated data is scarce. 

The architecture of the proposed SAM2-Adapter is illustrated in Figure \ref{framework}. We aim to keep the design of the adapter to be simple and efficient. Therefore, we choose to use an adapter that consists of only two MLPs and an activate function within two MLPs \cite{liu2023explicit}. It is worth noting that the different from SAM\cite{SAM}, the image encoder of SAM2 has four stages with hierarchical resolutions. Therefore, we initialized four different adapter and insert the four adapter in different layers of each stage. In each stage, the weight of the adapter is shared. Specifically, each of the adapter takes the information $F^i$ and obtains the prompt $P^i$: 
\begin{equation}
\label{get_context}
     P^i = {\rm MLP}_{up}\left({\rm GELU}\left({\rm MLP}_{tune}^i\left(F_i\right)\right)\right)
\end{equation}
in which ${\rm MLP}_{tune}^i$ are linear layers used to generate task-specific prompts for each Adapter. ${\rm MLP}_{up}$ is an up-projection layer shared across all Adapters that adjusts the dimensions of transformer features. $P^i$ refers to the output prompt that is attached to each transformer layer of SAM model. ${\rm GELU}$ is the GELU activation function \cite{hendrycks2016gaussian}. 
The information $F^i$ can be chosen to be in various forms. 

For more information, please refer to the original SAM-Adapter paper \cite{chen2023samfailssegmentanything}.

\subsection{Input Task-Specific Information}
It is worth noting that the information $F^i$ can be in various forms depending on the task and flexibly designed. For example, it can be extracted from the given samples of the specific dataset of the task in some form, such as texture or frequency information, or some hand-crafted rules. Moreover, the $F^i$ can be in a composition form consisting multiple guidance information:
\begin{align}
F_i & = \sum_{1}^{N}w_jF_j 
\end{align}
in which $F^j$ can be one specific type of knowledge/features and $w^j$ is an adjustable weight to control the composed strength. For more information, please refer to the original SAM-Adapter paper \cite{chen2023samfailssegmentanything}.

%% file: sec/4_Experiments.tex
\section{Experiments}
\subsection{Tasks and Datasets}
    In our experiments, we selected two challenging low-level structural segmentation tasks and one medical imaging task to evaluate the performance of the SAM2-Adapter: camouflaged object detection and shadow detection, and polyp segmentation.

\begin{figure*}[h]
\centering
\includegraphics[width=0.6\linewidth]{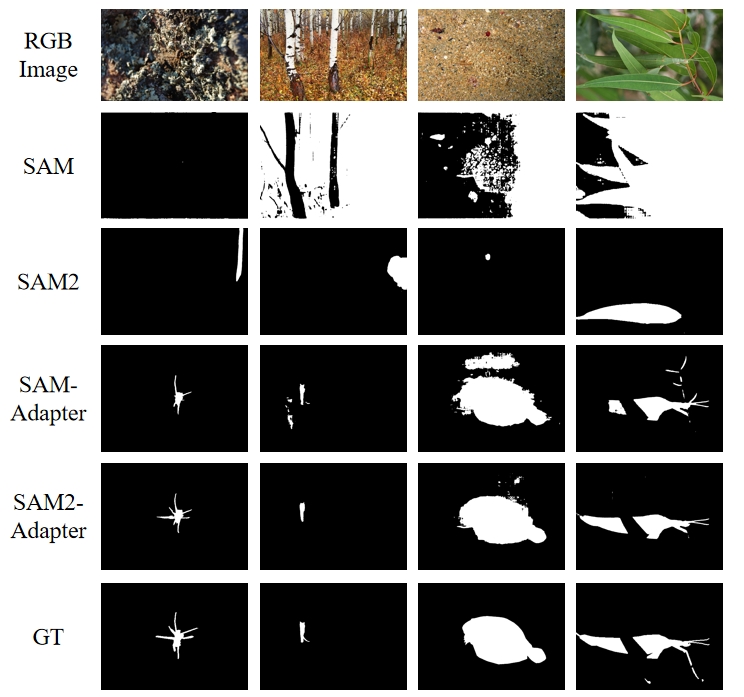}
\caption{\textbf{Visualization for Camouflaged Image Segmentation in CHAMELEON dataset} 
As shown in the figure, SAM often fails to detect animals that are visually camouflaged within their natural environments and can sometimes produce irrelevant results. SAM2 also struggles with similar issues and produces non-meaningful outcomes. However, by incorporating SAM-Adapter, our approach significantly improves object segmentation performance. Furthermore, SAM2-Adapter demonstrates even better performance than SAM-Adapter. The samples depicted are from the CHAMELEON dataset.} \label{fig2}
\end{figure*}

    For the camouflaged object detection task, we utilized three prominent datasets: COD10K \cite{fan2020camouflaged}, CHAMELEON \cite{skurowski2018animal}, and CAMO \cite{le2019anabranch}. COD10K is the largest dataset for camouflaged object detection, containing 3,040 training and 2,026 testing samples. CHAMELEON includes 76 images collected from the internet for testing. The CAMO dataset consists of 1,250 images, with 1,000 for training and 250 for testing. Following the training protocol in \cite{fan2020camouflaged}, we used the combined dataset of CAMO and the training set of COD10K for model training. For evaluation, we used the test sets of CAMO and COD10K, as well as the entire CHAMELEON dataset. For the shadow detection task, we employed the ISTD dataset \cite{wang2018stacked}, which contains 1,330 training images and 540 test images. For polyp segmentation (medical image segmentation), we use the kvasir-SEG dataset \cite{jha2020kvasir}. The train-test split followed the settings of the Medico multimedia task at MediaEval 2020: Automatic Polyp Segmentation \cite{jha2020medico}.

    For evaluation metrics, we followed the protocol in \cite{liu2023explicit} and used commonly-used metrics such as S-measure ($S_m$), mean E-measure ($E_\phi$), and MAE for the camouflaged object detection task. For the shadow detection task, we used the balance error rate (BER) metric. For the polyp segmentation task, we used mean Dice score (mDice) and mean Intersection-over-Union (mIoU) as the evaluation measures.

    For more details, please refer to the original SAM-Adapter paper \cite{chen2023samfailssegmentanything}.

\subsection{Implementation Details}
    In the experiment, we choose two types of visual knowledge, patch embedding $F_{pe}$ and high-frequency components $F_{hfc}$, following the same setting in \cite{liu2023explicit}, which has been demonstrated effective in various of vision tasks. $w^j$ is set to 1. Therefore, the $F_i$ is derived by $F_i=F_{hfc}+F_{pe}$.

\begin{table*}[t]
\centering
\scalebox{0.9}{
\begin{tabular}{c||cccc|cccc|cccc}
\hline
\multirow{2}{*}{Method} & \multicolumn{4}{c|}{CHAMELEON  \cite{skurowski2018animal}}                                    & \multicolumn{4}{c|}{CAMO \cite{le2019anabranch}}                                         & \multicolumn{4}{c}{COD10K \cite{fan2020camouflaged}}                                        \\ \cline{2-13} 
& $ S_\alpha \uparrow$               & $E_\phi \uparrow$              & $F^\omega_\beta \uparrow$              & MAE $\downarrow$           & $S_\alpha \uparrow$              & $E_\phi \uparrow$              & $F^\omega_\beta \uparrow$               & MAE $\downarrow$           & $S_\alpha \uparrow$              & $E_\phi \uparrow$              & $F^\omega_\beta \uparrow$               & MAE $\downarrow$           \\ \hline
SINet\cite{SINet}                   & 0.869          & 0.891          & 0.740          & 0.440          & 0.751          & 0.771          & 0.606          & 0.100          & 0.771          & 0.806          & 0.551          & 0.051          \\
RankNet\cite{RankNet}                 & 0.846          & 0.913          & 0.767          & 0.045          & 0.712          & 0.791          & 0.583          & 0.104          & 0.767          & 0.861          & 0.611          & 0.045          \\
JCOD \cite{JCOD}                   & 0.870          & 0.924          & -              & 0.039          & 0.792          & 0.839          & -              & 0.82           & 0.800          & 0.872          & -              & 0.041          \\
PFNet \cite{PFNet}                  & 0.882          & 0.942 & 0.810          & 0.330          & 0.782          & 0.852          & 0.695          & 0.085          & 0.800          & 0.868          & 0.660          & 0.040          \\
FBNet  \cite{FBNet}                 & 0.888          & 0.939          & 0.828 & 0.032 & 0.783          & 0.839          & 0.702          & 0.081          & 0.809          & 0.889          & 0.684          & 0.035          \\
\hline
SAM   \cite{SAM}    & 0.727 &  0.734   &  0.639    &  0.081  & 0.684 & 0.687 & 0.606 & 0.132  &  0.783    &    0.798     & 0.701 & 0.050   \\
SAM2   \cite{ravi2024sam2segmentimages}    & 0.359 &  0.375   &  0.115    &  0.357  & 0.350 & 0.411 & 0.079 & 0.311  &  0.429    &    0.505     & 0.115 & 0.218   \\
SAM-Adapter \cite{chen2023samfailssegmentanything,chen2023sam}                   & 0.896 & 0.919          & 0.824          & 0.033          & 0.847 & 0.873          & 0.765          & 0.070          & 0.883 & 0.918 & 0.801 & 0.025 \\
\hline
\textbf{SAM2-Adapter (Ours)}                    & \textbf{0.915} & \textbf{0.955}          & \textbf{0.889}          & \textbf{0.018}          & \textbf{0.855} & \textbf{0.909}          & \textbf{0.810}          & \textbf{0.051}          & \textbf{0.899} & \textbf{0.950} & \textbf{0.850} & \textbf{0.018} \\ \hline
\end{tabular}}
\caption{Quantitative Segmentation Result Comparison for Camouflaged Object Detection}
\label{Quantitative_Segmentation_Result}
\end{table*}

\begin{figure*}[h]
\centering
\includegraphics[width=0.6\linewidth]{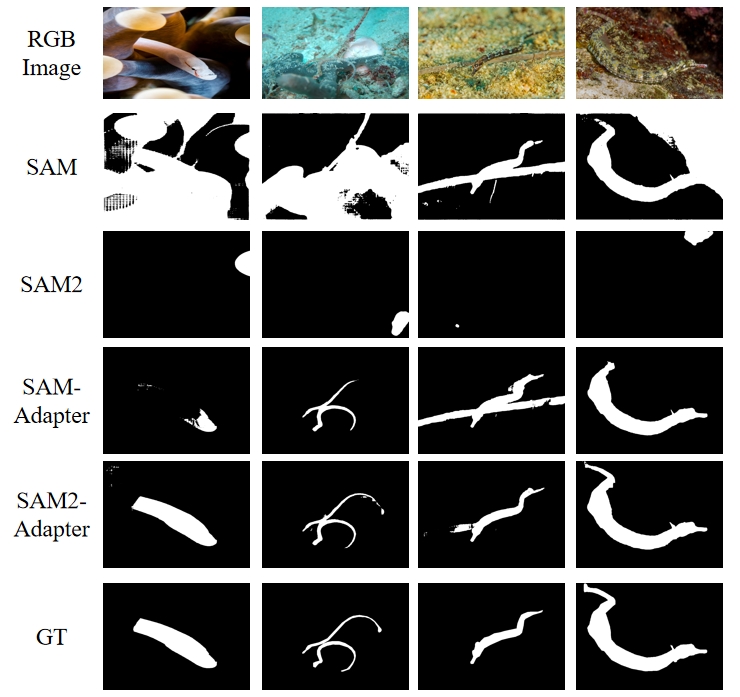}
\caption{\textbf{Camouflaged Image Segmentation in COD-10K dataset} As shown in the figure, SAM struggles to detect animals that are visually camouflaged within their natural environments and can sometimes produce results that lack meaningful segmentation. SAM2 also faces similar challenges, often resulting in no output or false results. However, by incorporating SAM2-Adapter, our method significantly improves object segmentation performance, surpassing SAM-Adapter. } \label{fig3}
\end{figure*}

\begin{figure*}[h]
\centering
\includegraphics[width=0.75\linewidth]{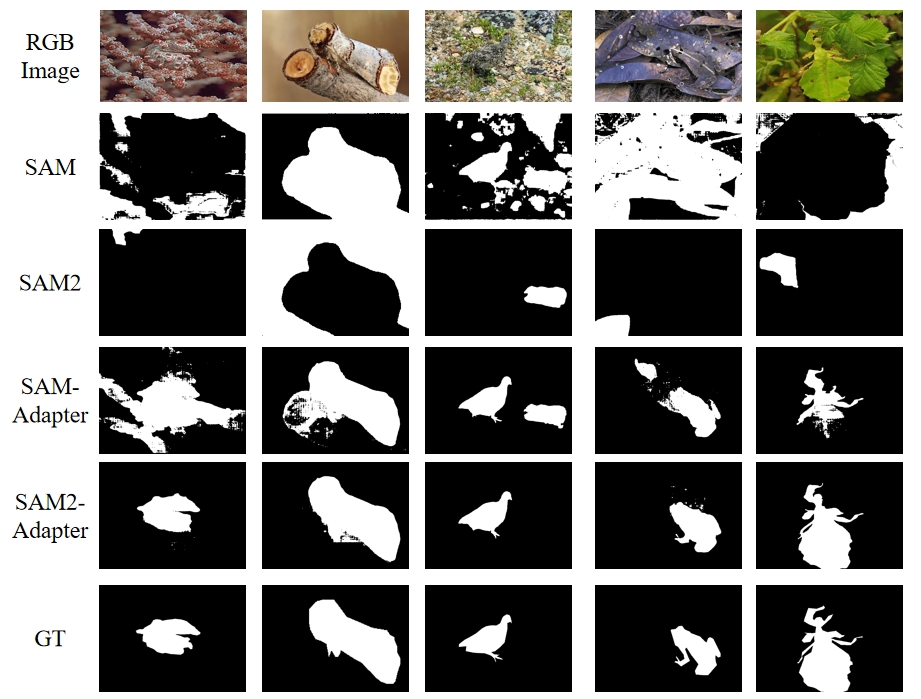}
\caption{\textbf{Camouflaged Segmentation of CAMO dataset.} The SAM and SAM 2 failed to perceive those animals that are visually ‘hidden’/concealed in their natural surroundings. The effectiveness of SAM2-Adapter is further validated in the visualized results.} \label{fig:4}
\end{figure*}

    The ${\rm MLP}_{tune}^i$ has one linear layer and ${\rm MLP}_{up}^i$ is one linear layer that maps the output from GELU activation to the number of inputs of the transformer layer. We use hiera-large version of SAM2. Balanced BCE loss is used for shadow detection. 
    
    BCE loss and IOU loss are used for camouflaged object detection and polyp segmentation. AdamW optimizer is used for all the experiments. The initial learning rate is set to 2e-4. Cosine decay is applied to the learning rate. The training of camouflaged object segmentation is performed for 20 epochs. Shadow segmentation is trained for 90 epochs. Polyp segmentation is trained for 20 epochs. The experiments are implemented using PyTorch on three NVIDIA Tesla A100 GPUs. For more information, please refer to the original SAM-Adapter paper \cite{chen2023samfailssegmentanything} and our codebase.

\subsection{Experiments for Camouflaged Object Detection}
    We first evaluated SAM on the challenging task of camouflaged object detection, where foreground objects often blend with visually similar background patterns. Our experiments revealed that SAM did not perform well in this task. As shown in Figure \ref{fig2}, SAM failed to detect several concealed objects. This was further confirmed by the quantitative results presented in Table \ref{Quantitative_Segmentation_Result}, where SAM's performance was significantly lower than existing state-of-the-art methods across all evaluated metrics, while SAM2, on its own, had the lowest performance, which fails to produce any meaningful results.

    In contrast, Figure \ref{fig2}, \ref{fig3} and \ref{fig:4} clearly demonstrates that by introducing the SAM2-Adapter, our method significantly elevates the model's performance. Our approach successfully identifies concealed objects, as evidenced by clear visual results. Quantitative results also show that our method outperforms the existing state-of-the-art methods.

    Furthermore, the SAM2-Adapter set a new SOTA performance. Visualized results show that SAM2-Adapter segments more precisely without adding extra false information, further demonstrating the robustness and accuracy of our approach. 

\subsection{Experiments for Shadow Detection}

\begin{figure*}[h]
\centering
\includegraphics[width=0.58\linewidth]{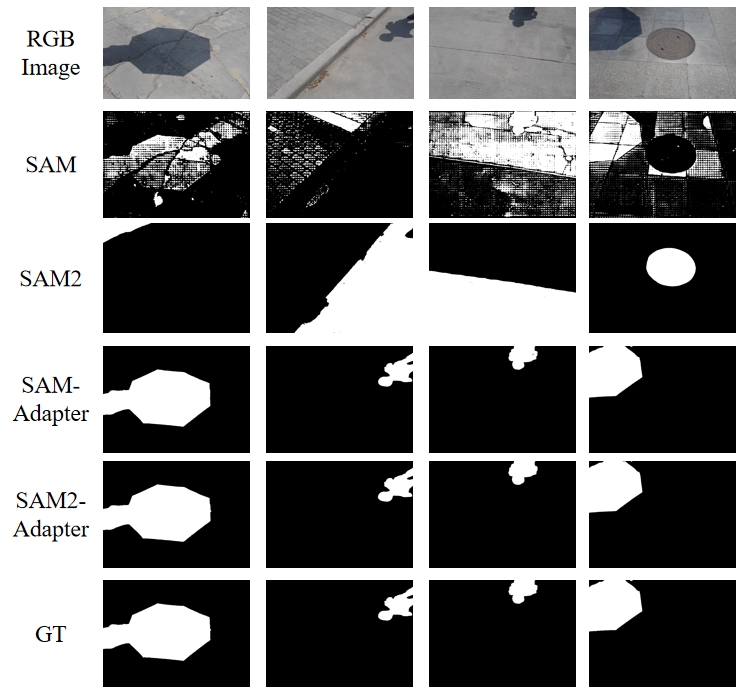}
\caption{\textbf{Shadow Detection Visualized.} Both SAM and SAM2 have no understanding of the "shadow" concept without proper prompting. They produce meaningless results. SAM-Adapter and SAM2-Adapter perform equally well in shadow detection tasks.} \label{fig5}
\end{figure*}

    We also evaluated SAM on shadow detection. However, as depicted in Figure \ref{fig5}, SAM struggled to differentiate between the shadow and the background, with parts missing or mistakenly added.

\begin{figure*}[h]
\centering
\includegraphics[width=0.57\linewidth]{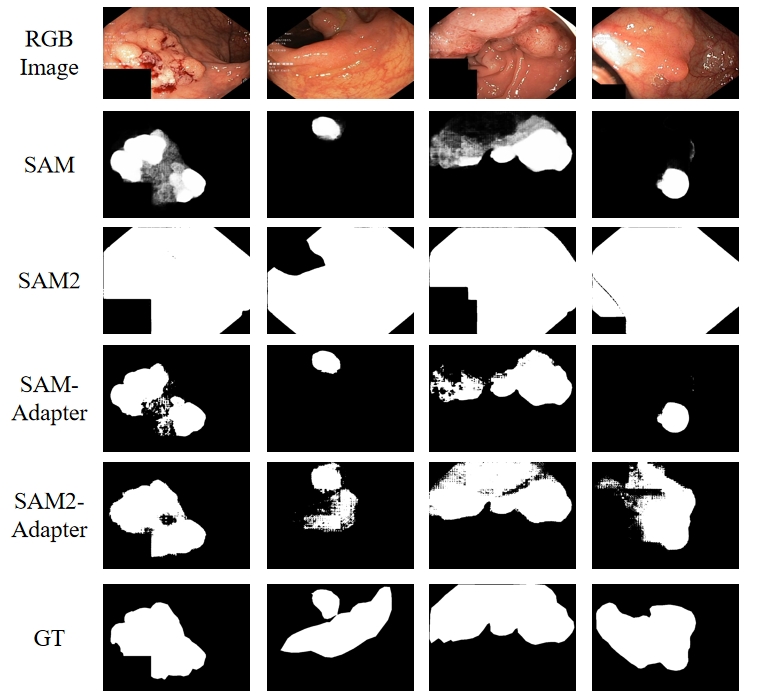}
\caption{\textbf{Visualization of Polyp Segmentation Results.} As illustrated in the figure, although SAM can identify some polyp structures in the image, the result is not accurate. Without proper prompting, SAM 2 failed to deliver meaningful polyp segmentation results. By using
SAM2-Adapter, our approach significantly outperforms SAM-Adapter with more accurate (and complete) segmentation results.} \label{fig:6}
\end{figure*}

\begin{table}[t]
    \centering  
    \begin{adjustbox}{width=0.25\textwidth,center}  
    \begin{tabular}{l | c}
    \toprule
    Method & BER $\downarrow$ \\
    \midrule
    Stacked CNN \cite{vicente2016large} & 8.60\\
    \midrule
    BDRAR \cite{BDRAR} & 2.69 \\
    \midrule
    DSC \cite{DSC} & 3.42\\
    \midrule
    DSD \cite{DSD} & 2.17 \\
    \midrule
    FDRNet \cite{zhu2021mitigating} & 1.55 \\
    \midrule
    SAM \cite{SAM} & 40.51\\
    SAM2 \cite{ravi2024sam2segmentimages} & 50.81\\
    SAM-Adapter  & \textbf{1.43}\\
    SAM2-Adapter (Ours) & \textbf{1.43}\\
    \bottomrule
    \end{tabular}
    \end{adjustbox}
    \caption{Result for Shadow Detection}
    \label{ablation_components}
\end{table}

Similarly, SAM2 also struggled with the "shadow" concept without proper prompting, failing to produce meaningful results. In our study, we compared various methods for shadow detection and found that SAM's performance was significantly poorer than existing methods. However, by integrating the SAM-Adapter, we achieved a substantial improvement in performance. The SAM-Adapter enhanced the detection of shadow regions, making them more clearly identifiable. Furthermore, SAM2-Adapter worked just as effectively as SAM-Adapter, delivering comparable results. Our findings were validated through quantitative analysis, and Table \ref{ablation_components} demonstrates the significant performance boost provided by the SAM-Adapter and matched by the SAM2-Adapter for shadow detection.

\subsection{Experiments for Polyp Segmentation}

\begin{table}{}{
    \begin{adjustbox}{width=0.67\columnwidth,center}
    \begin{tabular}{l | c c c}
    \toprule
    Method & mDice $\uparrow$ & mIoU $\uparrow$\\
    \midrule
    UNet \cite{unet} & 0.821 & 0.756\\
    \midrule
    UNet++ \cite{zhou2018unet++} & 0.824 & 0.753 \\
    \midrule
    SFA \cite{fang2019selective} & 0.725 & 0.619 \\
    \midrule
    SAM \cite{SAM} & 0.778 & 0.707\\
    SAM2 \cite{ravi2024sam2segmentimages} & 0.200 & 0.029\\
    SAM-Adapter  & 0.850 & 0.776\\
    SAM2-Adapter (Ours) & \textbf{0.873} & \textbf{0.806}\\
     \bottomrule
    \end{tabular}
    \end{adjustbox}
    \caption{Result for Polyp Segmentation}
    \label{fig6}}
\end{table}

    Following SAM-Adapter \cite{chen2023samfailssegmentanything}, we illustrate the application of SAM2-Adapter in the context of medical image segmentation, specifically focusing on polyp segmentation. Polyps, which have the potential to become malignant, are identified during colonoscopy and removed through polypectomy. Accurate and swift detection and removal of polyps are crucial in preventing colorectal cancer, a leading cause of cancer-related deaths globally.

\begin{table*}[h]
\centering
\scalebox{0.9}{
\begin{tabular}{c||cccc|cccc|cccc}
\hline
\multirow{2}{*}{Method} & \multicolumn{4}{c|}{CHAMELEON  \cite{skurowski2018animal}}                                    & \multicolumn{4}{c|}{CAMO \cite{le2019anabranch}}                                         & \multicolumn{4}{c}{COD10K \cite{fan2020camouflaged}}                                        \\ \cline{2-13} 
& $ S_\alpha \uparrow$               & $E_\phi \uparrow$              & $F^\omega_\beta \uparrow$              & MAE $\downarrow$           & $S_\alpha \uparrow$              & $E_\phi \uparrow$              & $F^\omega_\beta \uparrow$               & MAE $\downarrow$           & $S_\alpha \uparrow$              & $E_\phi \uparrow$              & $F^\omega_\beta \uparrow$               & MAE $\downarrow$           \\ \hline

SAM   \cite{SAM}    & 0.727 &  0.734   &  0.639    &  0.081  & 0.684 & 0.687 & 0.606 & 0.132  &  0.783    &    0.798     & 0.701 & 0.050   \\
SAM2   \cite{ravi2024sam2segmentimages}    & 0.359 &  0.375   &  0.115    &  0.357  & 0.350 & 0.411 & 0.079 & 0.311  &  0.429    &    0.505     & 0.115 & 0.218   \\
SAM-Adapter \cite{chen2023samfailssegmentanything,chen2023sam}                   & 0.896 & 0.919          & 0.824          & 0.033          & 0.847 & 0.873          & 0.765          & 0.070          & 0.883 & 0.918 & 0.801 & 0.025 \\
\hline
\textbf{SAM-Adapter with SAM2}                    & 0.909 &
0.930  &
0.820 &
0.031 &
0.853 &
0.870 &
0.747 &
0.071  &
0.884  &
0.916 &
0.774 &
0.028 \\
\hline
\textbf{SAM2-Adapter (Ours)}                    & \textbf{0.915} & \textbf{0.955}          & \textbf{0.889}          & \textbf{0.018}          & \textbf{0.855} & \textbf{0.909}          & \textbf{0.810}          & \textbf{0.051}          & \textbf{0.899} & \textbf{0.950} & \textbf{0.850} & \textbf{0.018} \\ \hline
\end{tabular}}
\caption{Quantitative Segmentation Result Comparison for Camouflaged Object Detection}
\label{abla}
\end{table*}
    While numerous deep-learning approaches have been developed for polyp identification, and the pre-trained SAM model shows promise in identifying some polyps, its performance can be significantly improved with our SAM-Adapter approach. However, without proper prompting, the SAM2 model fails to produce meaningful results. Our SAM2-Adapter addresses this issue and outperforms the original SAM-Adapter. The results of our study, presented in Table \ref{fig6} and the visualization results in Figure \ref{fig:6}, underscore the effectiveness of SAM2-Adapter in improving the accuracy and reliability of polyp detection.

\subsection{Ablation Study}
In this section, we further compare our adapter setting with that of the previous SAM-Adapter \cite{chen2023sam,chen2023samfailssegmentanything}. We aim to answer the question: 

\textit{Is the performance gain of SAM2-Adapter due to the improved backbone of SAM2, or the differences in the Adapter configurations? }

The answer is that it is a combination of both factors, which we show in our experiment with the results in Table. \ref{abla}. 

The key difference between the two approaches is in the Adapter configurations. The SAM-Adapter utilizes a single shared Adapter for all layers, while the SAM2-Adapter employs four distinct Adapters, each assigned to a different stage/resolution of the SAM2 model. 

In the ablation experiment, we use one shared adapter for SAM2 model while change the resolution with a single MLP layer to match the input, which we denoted as ``SAM-Adapter with SAM2".

By using multiple Adapters tailored to the hierarchical structure of SAM2, the SAM2-Adapter is able to more effectively capture and leverage the multi-resolution features learned by the large foundation model. This allows the Adapter to better adapt the general knowledge of SAM2 to the specific requirements of the downstream tasks.

In contrast, the single-Adapter approach of the original SAM-Adapter, while effective (achieved better performance than SAM-Adapter with a more powerful SAM2 backbone), may have been limited in its ability to fully harness the depth and nuances of the pre-trained model. The SAM2-Adapter's multi-Adapter configuration addresses this limitation, unlocking the full potential of the enhanced SAM2 backbone.

Therefore, the performance gains achieved by the SAM2-Adapter can be attributed to both the inherent improvements in the SAM2 model, as well as the more sophisticated Adapter architecture designed to better integrate the foundation model's capabilities with the target task requirements.

%% file: sec/5_Conclusion.tex
\section{Conclusion and Future Work}
    In this paper, we introduced SAM2-Adapter, a novel adaptation method designed to leverage the advanced capabilities of the Segment Anything 2 (SAM2) model for specific downstream segmentation tasks. Building on the success of the original SAM-Adapter, SAM2-Adapter utilizes a multi-adapter configuration that is specifically tailored to SAM2’s multi-resolution hierarchical Transformer architecture. This approach effectively addresses the limitations encountered with SAM, enabling the achievement of new state-of-the-art (SOTA) performance in challenging segmentation tasks such as camouflaged object detection, shadow detection, and polyp segmentation.

    Our experiments demonstrate that SAM2-Adapter not only retains the beneficial features of its predecessor, including generalizability and composability but also enhances these capabilities by integrating seamlessly with SAM2’s advanced architecture. This integration allows SAM2-Adapter to outperform previous methods and set new benchmarks across various datasets and tasks.

    The continued presence of challenges from SAM in SAM2 highlights the inherent complexities of applying foundation models to diverse real-world scenarios. Nevertheless, SAM2-Adapter effectively addresses these issues, showcasing its potential as a robust tool for high-quality segmentation in a range of applications.
    
    We encourage researchers and engineers to adopt SAM2 as the backbone for their segmentation tasks, coupled with SAM2-Adapter, to realize improved performance and advance the field of image segmentation. Our work not only extends the capabilities of SAM2 but also paves the way for future innovations in adapting large pre-trained models for specialized applications. Code, pre-trained models, and data processing protocols are available at \href{http://tianrun-chen.github.io/SAM-Adaptor/}{http://tianrun-chen.github.io/SAM-Adaptor/}